%% file: main.tex
\documentclass[journal]{IEEEtran}

\usepackage{color,array}

\usepackage[pdftex]{graphicx}
\DeclareGraphicsExtensions{.pdf,.jpeg,.png}

\usepackage[switch]{lineno} 
\usepackage{amsmath}
\usepackage{amssymb}
\usepackage{textcomp}
\usepackage{gensymb}
\usepackage{multirow}
\usepackage{bm}
\usepackage{graphicx}
\usepackage{tabularx}
\usepackage{xcolor}
\usepackage{soul}

\usepackage{booktabs}
\usepackage[ruled,vlined]{algorithm2e}

\newcommand{\etal}{\textit{et al}., }
\newcommand{\ie}{\textit{i}.\textit{e}., }
\newcommand{\eg}{\textit{e}.\textit{g}.\ }

\usepackage[colorlinks,urlcolor=blue,linkcolor=blue,citecolor=blue]{hyperref}

\usepackage[capitalize]{cleveref}

\usepackage{cite}

\begin{document}

\title{Self-Labeling in Multivariate Causality and Quantification for Adaptive Machine Learning}

\author{Yutian Ren, Aaron Haohua Yen, and  G. P. Li
% , \IEEEmembership{Fellow, IEEE}

\thanks{The authors are with the California Institute for Telecommunications
and Information Technology, University of California, Irvine, CA 92697
USA (e-mail: yutianr@uci.edu; ahyen@uci.edu; gpli@calit2.uci.edu).}
}

% \markboth{Journal of IEEE Transactions on xxx, Vol. 00, No. 0, Month 2020}
% {Yutian Ren \MakeLowercase{\textit{et al.}}: Bare Demo of IEEEtai.cls for IEEE Journals of IEEE Transactions on xxx}

\maketitle

\begin{abstract}
Adaptive machine learning (ML) aims to allow ML models to adapt to ever-changing environments with potential concept drift after model deployment. Traditionally, adaptive ML requires a new dataset to be manually labeled to tailor deployed models to altered data distributions.
Recently, an interactive causality based self-labeling method was proposed to autonomously associate causally related data streams for domain adaptation, showing promising results compared to traditional feature similarity-based semi-supervised learning. 
Several unanswered research questions remain, including self-labeling's compatibility with multivariate causality and the quantitative analysis of the auxiliary models used in the self-labeling. The auxiliary models, the interaction time model (ITM) and the effect state detector (ESD), are vital to the success of self-labeling. This paper further develops the self-labeling framework and its theoretical foundations to address these research questions. 
A framework for the application of self-labeling to multivariate causal graphs is proposed using four basic causal relationships, and the impact of non-ideal ITM and ESD performance is analyzed.
A simulated experiment is conducted based on a multivariate causal graph, validating the proposed theory.
\end{abstract}

\begin{IEEEkeywords}
Adaptive Learning, Self-Supervised Learning, Machine Learning, Causality Inspired Learning, Causal Time Delay, Noisy Label
% Enter key words or phrases in alphabetical order, separated by commas. For a list of suggested keywords, send a blank e-mail to \href{mailto:keywords@ieee.org}{\underline{keywords@ieee.org}} or visit \href{http://www.ieee.org/organizations/pubs/ani_prod/keywrd98.txt}{\underline{http://www.ieee.org/organizations/pubs/ani\_prod/keywrd98.txt}}
\end{IEEEkeywords}

\input{section1}

\input{section2}

\input{section3}

\input{section4}

\input{section5}

\input{section6}

\ifCLASSOPTIONcaptionsoff
  \newpage
\fi

\bibliographystyle{IEEEtran}
\bibliography{IEEEabrv,Bibliography}

\end{document}

%% file: section1.tex
\section{Introduction}
\label{sec:intro}

Supervised models form the majority of machine learning (ML) applications today due to their reliable performance despite requiring training dataset collection and annotation, consuming considerable time and labor \cite{label1, label2}. Adaptive ML allows models to adapt to environmental changes (\eg concept drift \cite{cd2019}) without full supervision to avoid laborious manual model adaptation. Several classes of methods have been proposed to achieve adaptive ML with minimum human intervention, including pseudo-labels, empowered by semi-supervised learning (SSL) \cite{yan2021augmented, slb_neurips}, delayed labels \cite{delay2, delayed_lb_review}, and domain knowledge enabled learning \cite{domainknowledge, stewart2017label}. 

Recently, self-labeling (SLB), a method based on interactive causality, has been proposed and demonstrated in \cite{ren2023slb} to equip AI models with the capability to adapt to concept drifts after deployment. The fundamental idea of self-labeling is to contextualize ML tasks with causal relationships, then apply the associated causation and learnable causal time lags (\ie interaction time) to causally related data streams, autonomously generating labels and selecting corresponding data segments that can be used as self-labeled datasets to adapt ML models to dynamic environments. It transforms complex problems on the cause side into easier problems on the effect side by temporally associating cause and effect data streams. Compared with traditional semi-supervised learning, self-labeling targets realistic scenarios with streaming data and is more theoretically sound for countering domain shifts without needing post-deployment manual data collection and annotation.

The self-labeling theory formulated in \cite{ren2023slb} leaves some key topics to be explored.
First, the proof and experiments in \cite{ren2023slb} use a minimal causal structure with two interacting variables. 
Causal graphs using Bayesian networks (\eg structural causal models \cite{pearl2009causality}) represent causality with four basic graph structures: chain, fork, collider, and confounder. The application of self-labeling in more complex causal graphs has not been well-defined. 
Second, the proof in \cite{ren2023slb} makes an implicit assumption that the auxiliary interaction time model (ITM) and effect state detector (ESD) are error-free with 100\% accuracy. In practical applications, however, ITM and ESD models are inaccurate, potentially degrading self-labeling performance. This needs extensive investigation to understand the impact of inaccuracy in the two auxiliary models \cite{noise1}. 
In addition, as self-labeling requires less manual annotation but more computing power for ITM and ESD inferencing, additional insights regarding the merit of self-labeling can be revealed by evaluating the tradeoffs between accuracy and cost. The cost herein includes the electricity consumed for compute \cite{label_cost2} and manpower cost for data annotation \cite{label_cost} and thus requires a shared metric for comparative evaluation.

This paper extends interactive causality enabled self-labeling theory and proposes solutions to these research questions. 
A domain knowledge modeling method is adopted using ontology and knowledge graphs with embedded causality among interacting nodes. 
This study explores the application of self-labeling to scenarios with multivariate causal structures via interaction time manipulation among multiple causal variables, focusing on the four basic causal structures extensible to more complex graphs. Additionally, we propose a method to quantify the impact of ITM and ESD inaccuracy on self-labeling performance using the dynamical systems (DS) theory and a metric incorporating the cost of human resources to evaluate tradeoffs along the spectrum of supervision.
A simulation utilizing a physics engine is conducted to demonstrate that self-labeling is applicable and effective in scenarios with complex causal graphs. 
It is also demonstrated experimentally that the interactive causality based self-labeling is robust to the uncertainty of ESD and ITM in practical applications. Self-labeling is also shown to be more cost-effective than fully supervised learning using a comprehensive metric.\footnote{Code will be available at \url{https://github.com/yutianRen/multi-cause-slb}.}

The structure of this paper is as follows. In \cref{sec:bg} we will briefly review self-labeling and its proof using dynamical system theory. \cref{sec: multi} develops self-labeling theory to applications with multivariate causal graphs, and \cref{sec: quant} addresses the questions of inaccurate ITM and ESD models and discusses the cost analysis of self-labeling. In \cref{sec: exp}, simulated experiments are illustrated and discussed to support the theory. We conclude this study in \cref{sec: end}.

%% file: section2.tex
\section{Background of Interactive Causality Enabled Self-Labeling}
\label{sec:bg}

In this section, we will briefly review the self-labeling method proposed by Ren \etal \cite{ren2023slb} to facilitate the reading of this article.

\subsection{Methodology}

The motivation of self-labeling originates from the necessity of domain adaptation to counter data distribution shifts (\eg concept drift) after ML models are deployed. To adapt a ML model (referred to as the task model) to the concept drift without the needs of manual data annotation, many types of methods have been proposed including unsupervised or semi-supervised domain adaptation \cite{uda1, Huang_2022_CVPR, Yu_2023_CVPR}, natural and delayed labels (such as user interactions in recommendation systems \cite{covington2016deep, natural_label, delay2, delayed_lb_review}), and domain knowledge based learning \cite{domainknowledge, stewart2017label}. 
Among them, the recent interactive causality enabled self-labeling is focused on automatic post-deployment dataset annotation by leveraging causal knowledge. In general, the data annotation consists of two steps in real applications: (1) select which samples to be labeled in streaming data; (2) generate labels for the selected samples. For static datasets, Step 1 is usually not required since the samples are selected already. The self-labeling method addresses the two steps by: (1) utilizing causality to find the sensor modalities that can generate labels for the task model; (2) inferring learnable causal time lags to associate labels from effects to the cause data to generate a dataset for retraining task models.

Self-labeling is applied to scenarios with interactive causality that represents an unambiguous causal relationship in an interaction between objects. Causality in general has various definitions across disciplines. The nomenclature of Interactive Causality is to emphasize that the causality leveraged for the self-labeling is associated with direct or indirect interactive activities among objects, which helps to identify useful causal relationships in application contexts for self-labeling.
Self-labeling leverages the temporal aspect of asynchronous causality, where interaction lengths and intervals are superimposed on time series data of sensing object states to form associations. 
In asynchronous causality, from the definition in physics, causes always precede effects, and the causal time lags between the occurrence of causes and effects is also referred to as the interaction time to emphasize the interactivity.
Self-labeling is predicated on the assumption that established causal relationships and the interaction time are less mutable than the input-output relations of ML models when there is concept drift, allowing self-labeling to adapt ML models to dynamic changes.

\begin{figure}[!t]
\centering
\includegraphics[width=0.9\columnwidth]{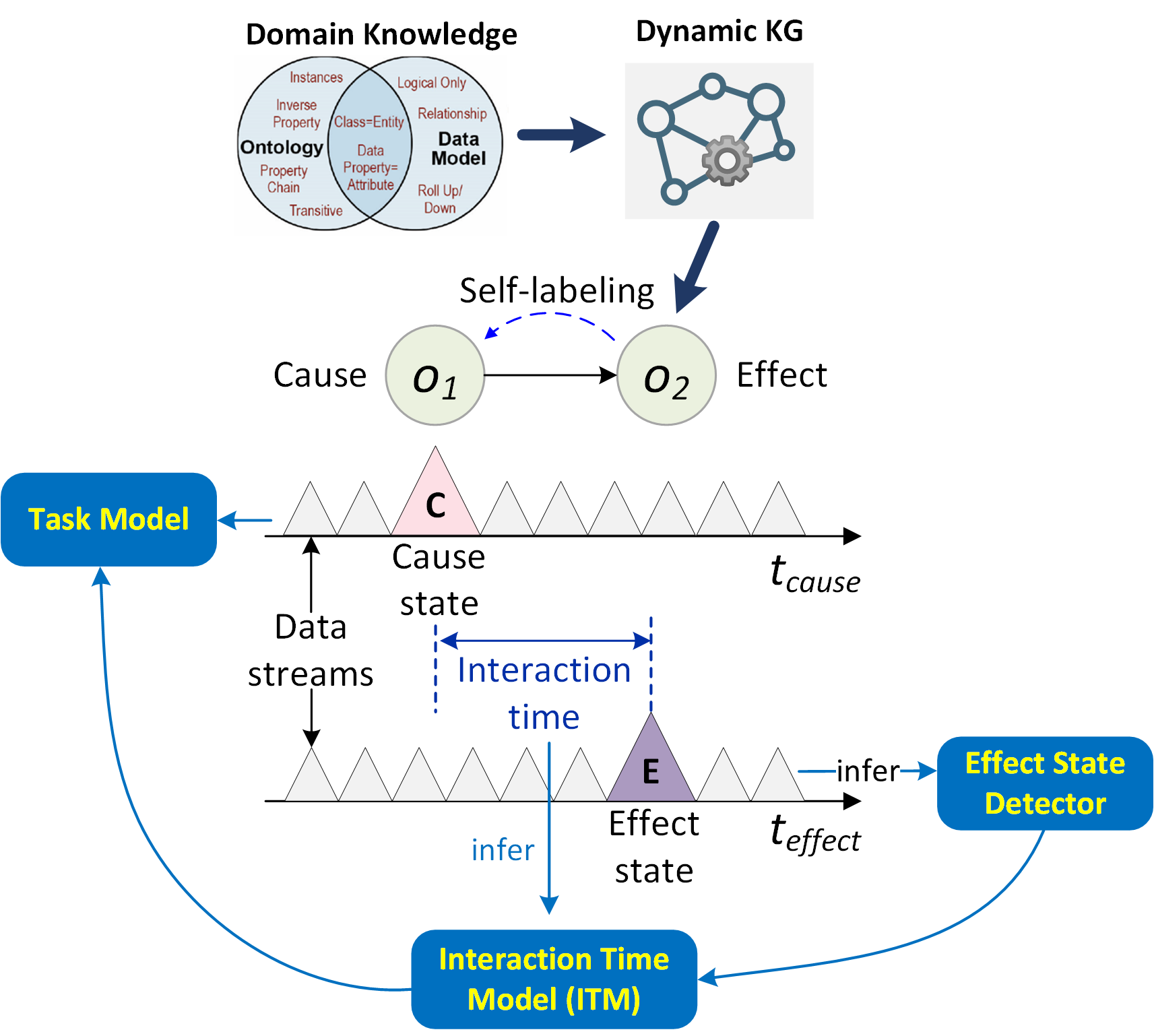}
\caption{An illustration of the overall procedure of self-labeling.}
\label{fig_kg}
\end{figure}

The self-labeling method works in the real-world environments with streaming data instead of static datasets. It captures and annotates samples from the real-time data streaming to generate a retraining dataset for task model domain adaptation because naturally data are acquired as streams in many real-world applications. 
Note that this does not mean the focus of self-labeling falls into the topics of time series domain adaptation \cite{ts1, ts2, ts3} which the self-labeling can be applied to.
The self-labeling method aims to assist ML tasks that are pattern recognition tasks accomplished by supervised machine learning models, which are referred to as task models.
The task model is the model for which the self-labeling provides automated procedural continual learning. 
Suppose an interaction scenario with two objects $o_{1}$ and $o_{2}$, as illustrated in \cref{fig_kg}, the causal relationship between objects $o_{1}$ and $o_{2}$ is known \textit{a priori} and extracted from the domain knowledge modeling. In this interaction context, the task model can be explained as ingesting the cause data related to $o_{1}$ to infer the effect on $o_{2}$. Due to the strong causal relationships, the effect state transitions indicate the state changes of the cause. Therefore, the effect can be utilized to generate labels for the task model. The effect state transitions are detected by the effect state detector (ESD) and constitute labels for training the task model. 
The temporal interval between cause and effect, defined as interaction time (\ie causal time lag \cite{timelag}), is utilized in self-labeling via prediction of the interaction time from effect data only, using a computational model known as the interaction time model (ITM.) The ITM is used to associate effects with the corresponding causes as training data, which accomplishes Step 1 of the data annotation procedure. 
% Task model inputs are selected by subtracting the effect state transition time from the ESD by the interaction time predicted by the ITM. 
Thus, inputs and labels are automatically generated for the continual learning of the task model, enabling adaptation to dynamic changes to input and/or output data distribution.
The intrinsic causality can be extracted from domain ontology \cite{Sawesi_Rashrash_Dammann_2022}, domain experts, documented knowledge (\eg standard operating procedure), or even knowledge distilled by large language models \cite{zhang2023understanding} and formulated as a dynamic causal knowledge graph (KG) \cite{ckg} with interactive nodes.

\subsection{Theory}

The proof of self-labeling in \cite{ren2023slb} uses a simplified dynamical system where two 1-$d$ systems $x$ and $y$ interact as 
\begin{align}
    & \dot{x} = f(x) +d(x) \label{df_nowind_a} \\ 
    & \dot{y} = y + h(x) \label{df_nowind_b}
\end{align}
where $f(\cdot)$ defines a vector field, $h(\cdot)$ is the coupling function, and $d(x)$ is the perturbation, simulating the impact of concept drift. Given an initial state ($x_{1}, y_{1}$) and final state ($x_{2}, y_{2}$), $x_{1}$ is defined as the cause state and $y_{2}$ as the effect state. The ML task is to learn a mapping between cause $x_{1}$ and effect $y_{2}$ in the $x$-$y$ interactive relationship.

We describe several key steps relevant to the scope of this paper from the full derivation in \cite{ren2023slb}. The derivation can be summarized in three steps: 1) in the original domain without perturbation, derive a relationship between inferred interaction time $t_{if}$ and $y_{2}$ that is to use effect $y_{2}$ to infer interaction time; 2) under perturbation derive the relation between $t_{if}$ and the self-labeled $x_{slb}$ that is to use $t_{if}$ to select corresponding $x$ as the self-labeled cause state; 3) cancel out $t_{if}$ to derive the relation between $y_{2}$ and $x_{slb}$ that is the learned task model by the self-labeling method. The intermediate steps of the above Step 1 and 2 are summarized as
\begin{align}
    & y_{2}=e^{t_{if}}\int_{0}^{t_{if}} e^{-\tau}\cdot h(A^{-1}(\tau+A_{x_{2}}-t_{if})) \,d\tau + e^{t_{if}}y_{1} \label{flow_wind_b} \\
    & {t_{if}} = B_{x_{2}}-B_{x_{slb}}. \label{tif_xslb}
\end{align}

The self-labeling method is compared with fully supervised (FS) and conventional semi-supervised (SSL) methods by solving the DS to derive 
\begin{multline}
    y_{2slb} = e^{B_{x_{2}}-B_{x_{slb}}} \cdot
    (\int_{0}^{B_{x_{2}}-B_{x_{slb}}} e^{-\tau}\\ h(A^{-1}(\tau+A_{x_{2}}-B_{x_{2}}+B_{x_{slb}})) \,d\tau + y_{1}). \label{y2slb} 
\end{multline}
\begin{multline}
    y_{2trad}=e^{A_{x_{2}}-A_{x_{1}}} \cdot
    (\int_{0}^{A_{x_{2}}-A_{x_{1}}} e^{-\tau} \\
    h(A^{-1}(\tau+A_{x_{1}})) \,d\tau 
    + y_{1}) \label{y2_trad}
\end{multline}
\begin{multline}
    y_{2fs}=e^{B_{x_{2}}-B_{x_{1}}} \cdot
    (\int_{0}^{B_{x_{2}}-B_{x_{1}}} e^{-\tau} \\
    h(B^{-1}(\tau+B_{x_{1}})) \,d\tau 
    + y_{1}) \label{y2_gt}
\end{multline}
where subscripts $slb$, $fs$, and $trad$ represent self-labeling, FS, and traditional SSL methods respectively. $A(x)=\int ^{x} \frac{1}{f(\xi)}\, d\xi$ and $B(x)=\int ^{x} \frac{1}{f(\xi)+d(\xi)}\, d\xi$. We use subscript $A_{x_{1}}$ to represent $A(x_{1})$ and so as $B(x)$.

Given the background, this study will extend the self-labeling theory to multivariate causality and comprehend the research questions proposed in \cref{sec:intro}.

%% file: section3.tex
\section{Self-Labeling in Multivariate Causal Graph}
\label{sec: multi}

\begin{figure}[!t]
\centering
\includegraphics[width=1.0\columnwidth]{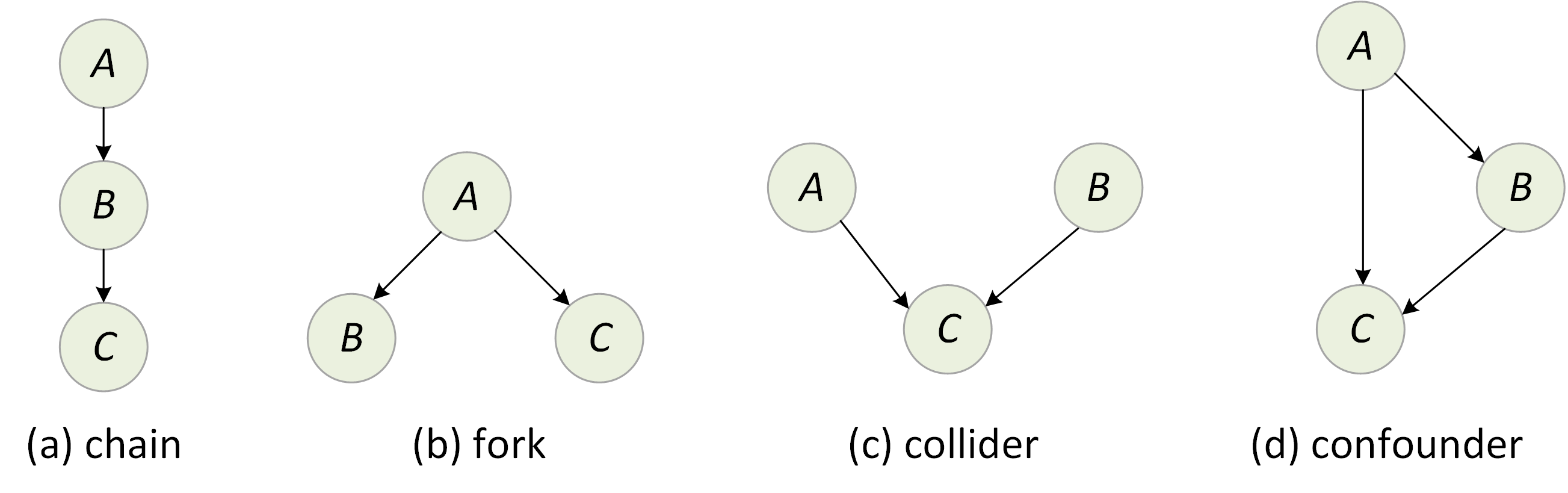}
\caption{Four basic causal structures represented in graphical models.}
\label{fig_4causal}
\end{figure}

The self-labeling is established on existing causal relationships. With more complex causal systems, causal graphs become an effective tool to represent the relations. 
In \cref{fig_kg}, a self-labeling scenario on a simple single cause and effect causal structure is illustrated as a foundation. 
In structural causal models, there are four basic causal structures, namely chain, collider, fork, and confounder \cite{pearl2009causality}, illustrated in \cref{fig_4causal} where nodes are variables and edges describe a forward causal relationship. We explore the application of self-labeling in these basic causal structures and their extension to more complex causal graphs.

Chains are a sequence of nodes forming a direct path from causes to effects. In the minimal example shown in \cref{fig_4causal}, $B$ acts as a mediator, facilitating the indirect influence of $A$ on $C$. Forks occur when a single cause produces multiple effects where $A$ is the common cause for effect $B$ and $C$. Colliders are situations where multiple causes ($A$ and $B$) converge to produce a single outcome $C$. A confounder is a third variable $A$ that affects both the cause $B$ and the effect $C$, making it difficult to establish a direct causal relationship between $B$ and $C$. 
Confounders complicate the causal analysis and causal effect inference. 
These four structures form the basis for most causal relationships and thus will be discussed in this section for their applications in self-labeling. 

An emerging question when a causal relationship involves multiple variables is how to organize and leverage the relationships, including interaction time, of each set of variables for self-labeling. Additionally, the undetermined logical relation among variables (\eg AND/OR/XOR) further complicates the relational analysis for self-labeling. 
The logical relations referred to here are the function space that maps cause variables to effect variables, \eg different logical relations of $A$ and $B$ in a collider to generate effect $C$. 
The following analysis of interaction time calculation does not assume specific logical relations, and focuses on the state transitions to maximize information available to the ITM.

Given a chain structure, the combined interaction time from $C$ to $A$ can be represented as
\begin{align}
    & t_{CA} = t_{CB} + t_{BA} \label{ch2_t_chain}
\end{align}
following the sequentially transmitted causal effect. Hence, the interaction time of two pairs of variables in a chain can be directly combined for the self-labeling between $A$ and $C$. 
Self-labeling here is invariant to the causal logical relations among the variables as the causal effect is passed independently.

\begin{figure}[!t]
\centering
\includegraphics[width=0.7\columnwidth]{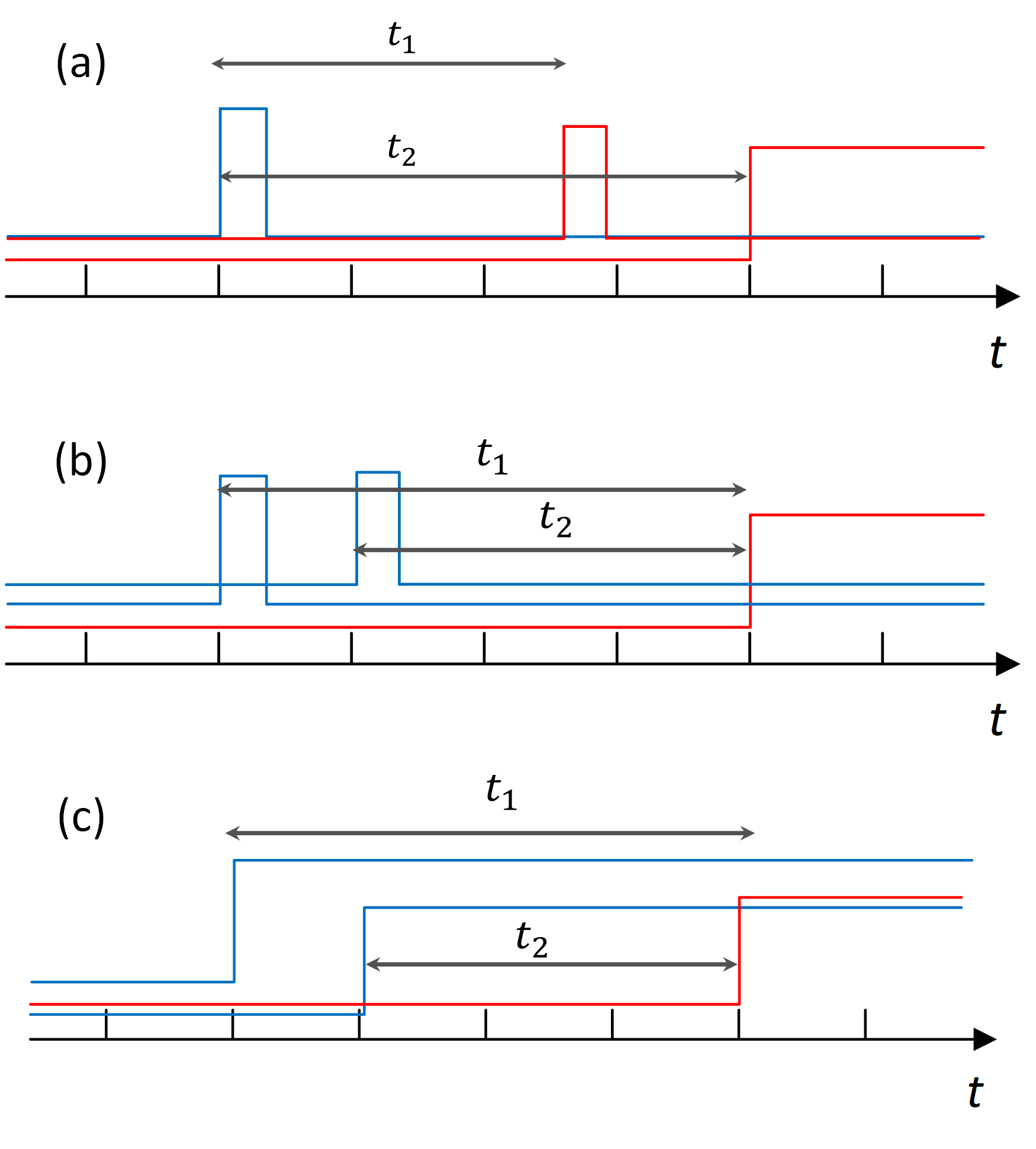}
\caption{An illustration of interaction time combination with (a) multiple effects, (b) multiple transient state causes, and (c) multiple steady state causes. Blue signals represent causes and red signals effects. $t_{1}$ and $t_{2}$ represent the interaction time of each cause-effect pair respectively.}
\label{fig_timing}
\end{figure}

In a fork structure, the multiple effects can individually or jointly label the cause depending on the availability of effect observers. The causal logical relations can limit the effectiveness of a subset or singular variable due to partial observability. In \cref{fig_4causal}(b), the necessary interaction time is represented as 
\begin{align}
    & t_{AC} = max(t_{AB}, t_{AC}) . \label{ch2_t_fork}
\end{align}
The combination of individual interaction times uses $max$ to capture all effect transitions for self-labeling. \cref{fig_timing}(a) depicts an example of a fork with both steady and transient effect states where the combined interaction time is $max(t_{1}, t_{2})$. The ESD and ITM process relevant portions of signals from $B$ and $C$ to infer the interaction time and label of $A$. In practice, observing additional effect channels can improve label granularity. 

In a collider, multiple cause variables jointly influence an effect variable. Regardless of the logical relations, the cause state changes can be defined between steady states or as transient states as shown in \cref{fig_timing}(b) and (c). State transitions between steady states as in \cref{fig_timing}(c) are less sensitive to ITM inaccuracies as state information persists after the transient portion of the cause signals.
In transient state to capture the information-rich state transitions of each cause variable, an individual ITM is used for each cause-effect pair to extract self-labeling input data. To preserve rich information between the involved relationships, the self-labeled data segments from multiple cause variables can form the input for the task model as a multi-dimensional input, allowing the task model to find discriminative features to learn the relation of $A$ and $B$ with regards to their joint effect $C$. Similar learning strategies can be adopted for other logical relations.

In the confounding structure, the confounder $A$ affects $B$ and $C$. If the self-labeled variable pair is $B$ and $C$, the confounder $A$ functions as an additional cause, which can be treated similarly to a collider. 
For the self-labeled pair $A$ and $C$, $B$ forms an intermediate cause and indirect path to the effect. 
In this case, the effect in $C$ can result from either path, each with distinct interaction times to $A$. To select the proper interaction time, $B$ must be observed to determine the causal path. However, in practice, a single ITM can be designed to infer the interaction time for $A$ and $C$ through either path by teaching the ITM differentiable characteristics of the two paths, which will be experimentally validated in \cref{sec: exp}.

In a more complex causal graph, the self-labeling schema for the four basic cases can be used as a tool to analyze the interaction time calculus by disentangling a complex graph into the four basic structures.

%% file: section4.tex
\section{Quantitative Analysis of Self-Labeling}
\label{sec: quant}

This section provides a comprehensive analysis of the ITM and ESD and a comparative cost analysis for self-labeling.

\subsection{Inaccuracy of ITM and ESD}

In practice, the ITM and ESD used in self-labeling are computational models with inherent inaccuracies. These inaccuracies can result in improper task model training inputs, shifting the learning away from the ground truth. In this section, we study the impact of ITM and ESD inaccuracy on self-labeling performance. 

The quantification of ITM inaccuracy is accomplished by imposing an error factor $\xi_{t}$ on the inferred interaction time $t_{if}=G(y_{2})$ in the self-labeling derivation outlined in \cref{sec:bg}, where $G(\cdot)$ represents the inverse function of \cref{flow_wind_b}. 
We define $\xi_{t}=t_{if}^{\xi_{t}} / t_{if}$ where $t_{if}$ is the error-free inferred interaction time and the error-imposed inferred interaction time is $ t_{if}^{\xi_{t}} = \xi_{t}G(y_{2})$, where $\xi_{t}=1.1$ represents a positive 10\% error and $\xi_{t}=0.9$ a negative 10\% error. The learned $y_{2slb}$ and $x_{slb}$ relation upon a $\xi_{t}$ inaccurate ITM is
\begin{multline}
    y_{2slb}^{\xi_{t}} = e^{\frac{1}{\xi_{t}}(B_{x_{2}}-B_{x_{slb}})} \cdot
    (\int_{0}^{\frac{1}{\xi_{t}}(B_{x_{2}}-B_{x_{slb}})} e^{-\tau}\\ h(A^{-1}(\tau+A_{x_{2}}- \frac{1}{\xi_{t}} (B_{x_{2}}-B_{x_{slb}}))) \,d\tau + y_{1}). \label{y2slb_itm} 
\end{multline}

To analyze the ITM error's impact on self-labeling, we find the derivative of \cref{y2slb_itm} to be
\begin{align}
    \frac{dy_{2slb}^{\xi_{t}}}{dx_{slb}}&=-\frac{1}{\xi_{t}}B^{'}_{x_{slb}}e^{\frac{1}{\xi_{t}}(B_{x_{2}}-B_{x_{slb}})} (y_{1}+ \nonumber\\ & h(A^{-1}(A_{x_{2}}- \frac{1}{\xi_{t}} (B_{x_{2}}-B_{x_{slb}})))). \label{dy_slb}
\end{align}
It is challenging to analytically derive the impact of $\xi_{t}$ in \cref{dy_slb}. For specific scenarios with numerical representations, the impact of $\xi_{t}$ can be analyzed accordingly. We will discuss a numerical example in the next section.

An error factor $\xi_{e}$ is introduced to quantify the impact of ESD inaccuracy. With both $\xi_{t}$ and $\xi_{e}$, \cref{y2slb} becomes
\begin{multline}
    y_{2slb}^{\xi} = \frac{1}{\xi_{e}} e^{\frac{1}{\xi_{t}}(B_{x_{2}}-B_{x_{slb}})} \cdot
    (\int_{0}^{\frac{1}{\xi_{t}}(B_{x_{2}}-B_{x_{slb}})} e^{-\tau}\\ h(A^{-1}(\tau+A_{x_{2}}-\frac{1}{\xi_{t}}(B_{x_{2}}-B_{x_{slb}}))) \,d\tau + y_{1}). \label{y2slb_itm_esd} 
\end{multline}
Likewise, the derivative of \cref{y2slb_itm_esd} can be used to quantify the impact analytically. A numeral example is provided in \cref{sec:ds_ex}.

Note the conceptual difference between the ground truth interaction time $t_{true}$, error-free inference $t_{if}$, and error-imposed $t_{if}^{\xi_{t}}$. This study focuses on ITM model inaccuracy ($t_{if}$ versus $t_{if}^{\xi_{t}}$), rather than cases where $t_{if}$ is unequal to $t_{true}$ due to the perturbation.

\subsection{DS examples for ITM and ESD quantification.}
\label{sec:ds_ex}

We will use a numerical example of a dynamical system to discuss the impact of ITM and ESD inaccuracy on self-labeling performance. Given $f(x)=x$, $d(x)=x$, $\xi_{t}$, and $\xi_{e}$, we can solve \cref{df_nowind_a} and \cref{df_nowind_b} and derive
\begin{align}
    y_{2slb} = \frac{1}{\xi_{e}} (x_{2} log(x_{2}/x_{1})^{\frac{1}{2\xi_{t}}} + y_{1} (x_{2}/x_{1})^{\frac{1}{2\xi_{t}}}). \label{ds_itm_esd}
\end{align}

Based on \cref{y2_gt}, the fully supervised equivalent is
\begin{equation}
    y_{2fs} = x_2 - (x_1x_2)^{\frac{1}{2}} + y_1(x_2/x_1)^{\frac{1}{2}}. \label{ds_yfs}
\end{equation}

\begin{figure}[!t]
\centering
\includegraphics[width=0.7\columnwidth]{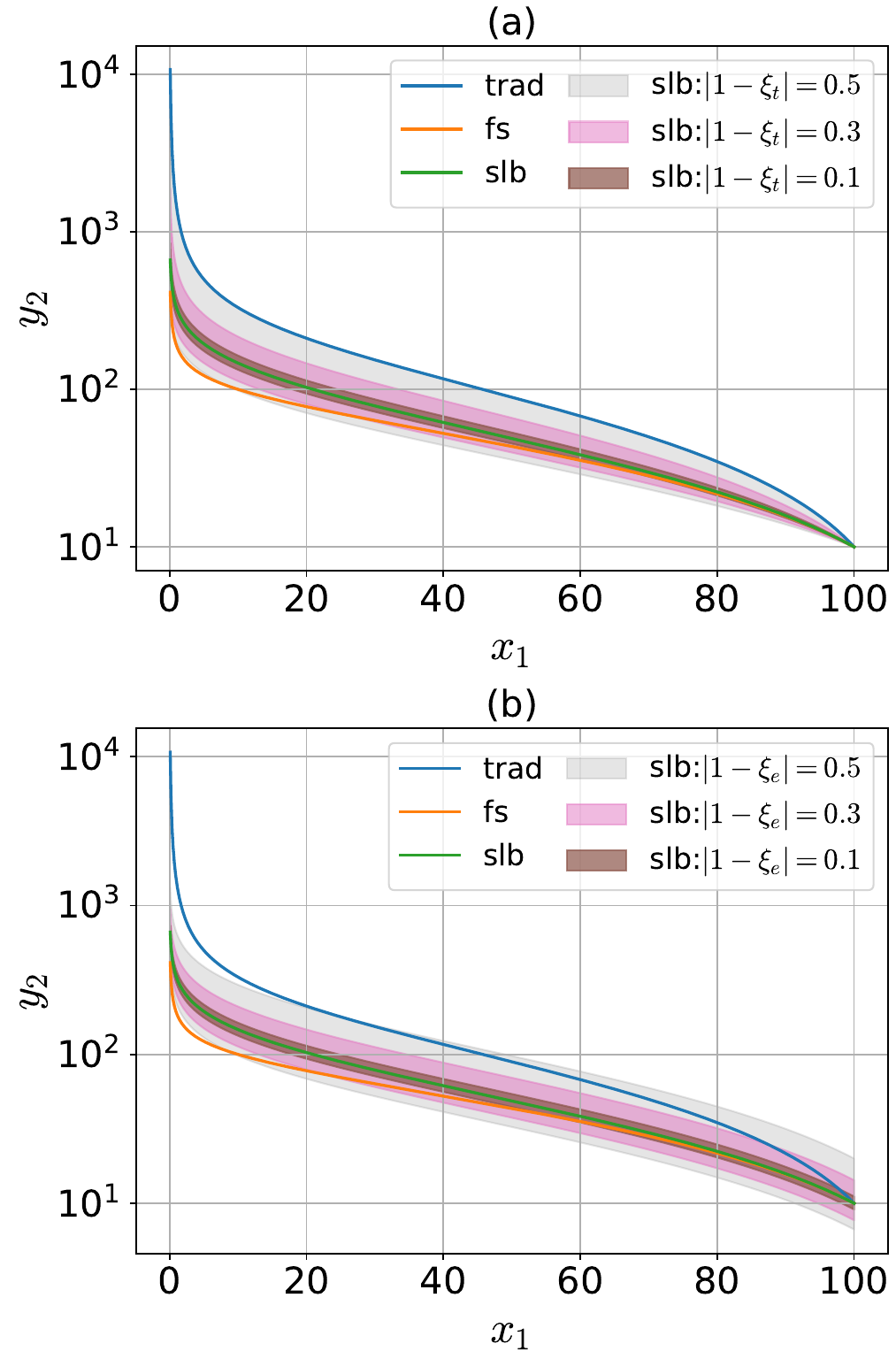}
\caption{DS example with (a) ITM and (b) ESD errors. The error bounds are represented by colored regions. $y_{2}$ axis is in $log$ scale.}
\label{fig_ds_itm_esd}
\end{figure}

In the numerical self-labeled example above, we can adjust the error factors to visualize ITM and ESD accuracy influence on the learning result. The result is shown in \cref{fig_ds_itm_esd} where $\xi_{t}$ and $\xi_{e}$ are changed independently. In \cref{fig_ds_itm_esd}, it is evident that self-labeling holds strong potential in ITM and ESD error tolerance given its performance at a 30\% error ratio, indicating that in practical applications with non-ideal ITM and ESD models, self-labeling can still retain its overwhelming performance advantage over traditional SSL methods.

\subsection{ITM as a Sampling Window.}

In the self-labeling procedure, the ITM infers the interaction time from effect state change to cause state change. This defines a sampling window on the cause data stream, selecting the relevant data segments. As the sampled data is directly used to train the task model, it is necessary for the ITM to maintain the desired sampling behavior, as described in this section.

The ITM accuracy is necessarily bound by an acceptable error margin $\epsilon$, defined as the deviation of the SLB-learned $y_{2slb}$ from the optimal learning result $y_{2fs}$ for the same $x_{1}$. The acceptable bounds for $y_2$, $y_{2low}$ and $y_{2high}$, can be expressed as $(1-\epsilon) y_{2fs} \leq y_{2fs} \leq (1+\epsilon) y_{2fs}$ in relation to the defined error margin.
Substituting $y_{2slb}$ in \cref{flow_wind_b} with $y_{2low}$ and $y_{2high}$ provides similar bounds $t_{iflow}$ and $t_{ifhigh}$ for $t_{if}$. For comparison, the actual inferred interaction time and corresponding $y_{2}$ are calculated following the regular derivation procedures in \cref{tif_xslb} and \cref{y2slb} and compared with the bounds for $y_{2}$ and $t_{if}$.

Using the numerical example in \cref{sec:ds_ex} with $x_{1}=80$, we can substitute $x_{1}$ in \cref{ds_yfs}
to obtain $y_{2fs}=21.7376$, the optimal learning result at $x_{1}=80$. For $\epsilon=0.5$, $y_{2high}=32.6064$ and $y_{2low}=10.8688$, substitution of $y_{2slb}$ by $y_{2high}$ and $y_{2low}$ in $y_{2slb}=x_{2} t_{if} + y_{1} e^{t_{if}}$ produces $t_{ifhigh}=0.2035$ and $t_{iflow}=0.0079$. 

The nominal $t_{if}$ and $y_{2slb}$ can be derived from $t_{if}=log\sqrt{\frac{x_{2}}{x_{1}}}=0.11157$ and $y_{2slb}=x_{2}t_{if}+y_{1}e^{t_{if}}=22.3373$. 
The values $y_{2slb}=22.3373$ and $t_{if}=0.11157$ are within their respective error bounds, indicating that the current ITM sampler is satisfactorily accurate given $\epsilon=0.5$.

\begin{figure}[!t]
\centering
\includegraphics[width=0.6\columnwidth]{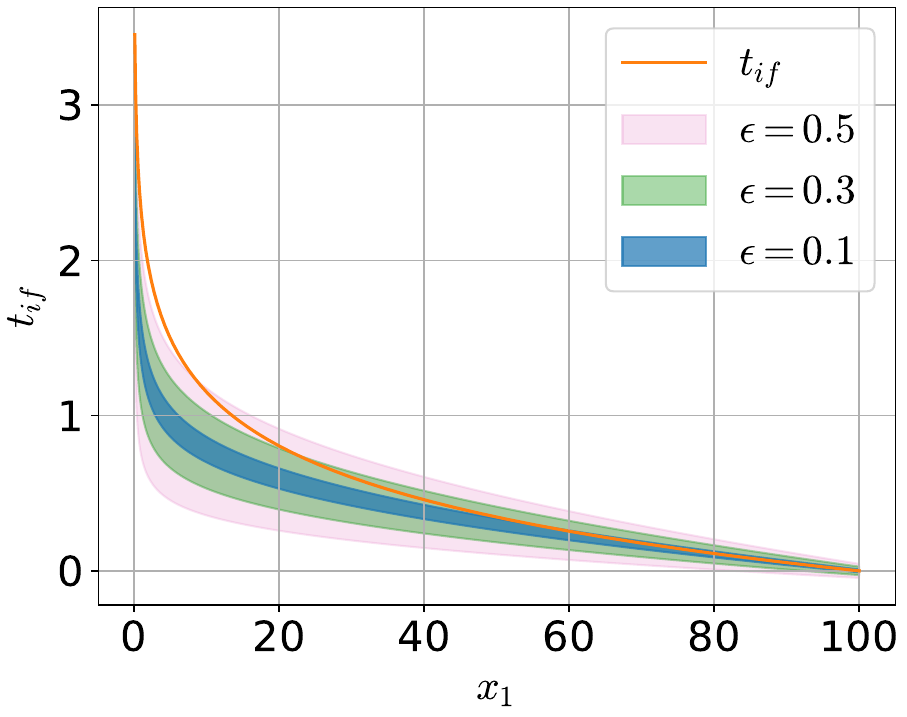}
\caption{A dynamical system example of the ITM as a data sampler with 10\%, 30\%, and 50\% error margins. The error bounds are represented by colored regions.}
\label{fig_ds_itm_sampler}
\end{figure}

The ITM error bound is visualized in \cref{fig_ds_itm_sampler}. It can be observed that with higher error tolerance $\epsilon$, the ITM requirement is relaxed, increasing the amount of allowable data samples.

\subsection{Cost Assessment of Self-Labeling}
\label{sec: cost}

The cost of producing an ML model arises not only from electricity consumed for computation but also from the human resources required for data labeling. This section proposes a metric to evaluate the cost and accuracy tradeoff between self-labeled, fully supervised, and semi-supervised learning.

To incorporate the cost of both labor and electricity, a cost index is defined to quantify the additional post-deployment cost for countering concept drift as
\begin{equation}
    cost\_index = \frac{\Delta acc}{E + M} \label{eq_cost_idx}
\end{equation}
where $\Delta acc$ is accuracy variation after deployment, $E$ represents the electricity cost, and $M$ represents the cost of manual labeling. $E$ and $M$ use the US dollar (\$) as their unit and can be represented as a product of the number of data samples $n$ and the unit cost per sample $C_{e}$ and $C_{m}$ respectively, 
\begin{align}
    E &= n \times C_{e} \label{eq_e_cost} \\ 
    M &= n \times C_{m}    \label{eq_m_cost}
\end{align}
$C_{m}$ is the labor cost to label a data sample, and $C_{e}$, the unit cost of electricity consumption, is estimated by using a product of needed compute time per sample $t_{compute}$, used GPU power $P$, and the electricity rate $r$ as
\begin{equation}
    C_{e} = t_{compute} (h) \times P (kW) \times r (\$) \label{eq_ce}
\end{equation}

In the post-deployment stage, self-labeled, fully supervised, and semi-supervised learning each involves differing operations, incurring labor and electricity costs. For self-labeling, post-deployment operations are ITM inference, ESD inference, and retraining on self-labeled datasets. Fully supervised learning requires manual dataset labeling and retraining on the newly labeled data to achieve continual learning. Semi-supervised learning's post-deployment costs are from continual training exclusively. To simplify the analysis, two assumptions are made: (1) the task model, ITM, and ESD use ML models with equal energy consumption; (2) the energy consumption from continuous ESD inference during periods with no state changes that are candidates for self-labeling is insignificant.
% as effect state classification is not required.

In the pre-deployment stage, the aggregate costs for FS and SSL are identical, namely the labeling and training of a pre-training dataset. SLB incurs additional costs from the labeling of effect state changes and interaction times and the training of ITM and ESD. Hence, the pre-deployment cost of a SLB system is higher than that of FS systems.

Additionally, a coefficient $\alpha$ is introduced to quantify the ratio between the duration needed for training and inference per sample where $\alpha \times C_{e_{train}}=C_{e_{infer}}$ as it is approximated that ITM, ESD, task model are equivalent in energy consumption but operate in different modes during self-labeling. 

For $cost\_index_{slb}$ to be greater than $cost\_index_{fs}$, the condition
\begin{equation}
    \frac{\Delta acc_{slb}}{E_{slb} + M_{slb}} \geq \frac{\Delta acc_{fs}}{E_{fs} + M_{fs}} \label{eq_acc}
\end{equation}
must be satisfied, resulting in 
\begin{equation}
    \frac{\Delta acc_{slb}}{\Delta acc_{fs}} \geq \frac{(1+2\alpha)\times t_{compute} \times P \times r}{t_{compute} \times P \times r + C_{m}} \times \beta \label{eq_acc3}
\end{equation}
where $\beta$ is the ratio of the number of data samples $n_{slb}$ and $n_{fs}$. 
\cref{sec: exp} includes case studies demonstrating these metrics. Note that this analysis is practically meaningful only when both FS and SLB systems can achieve and maintain the desired accuracy despite potential domain shifts after deployment.

%% file: section5.tex
\section{Experiment Results and Discussion}
\label{sec: exp}
This section provides a simulated experiment to demonstrate the self-labeling method for adaptive ML with complex causal structures and to quantify the impact of ITM and ESD uncertainties experimentally.

\subsection{Multi-cause simulation and self-labeling experiment}
\label{sec: exp_mul}
\begin{figure}[!t]
\centering
\includegraphics[width=1.0\columnwidth]{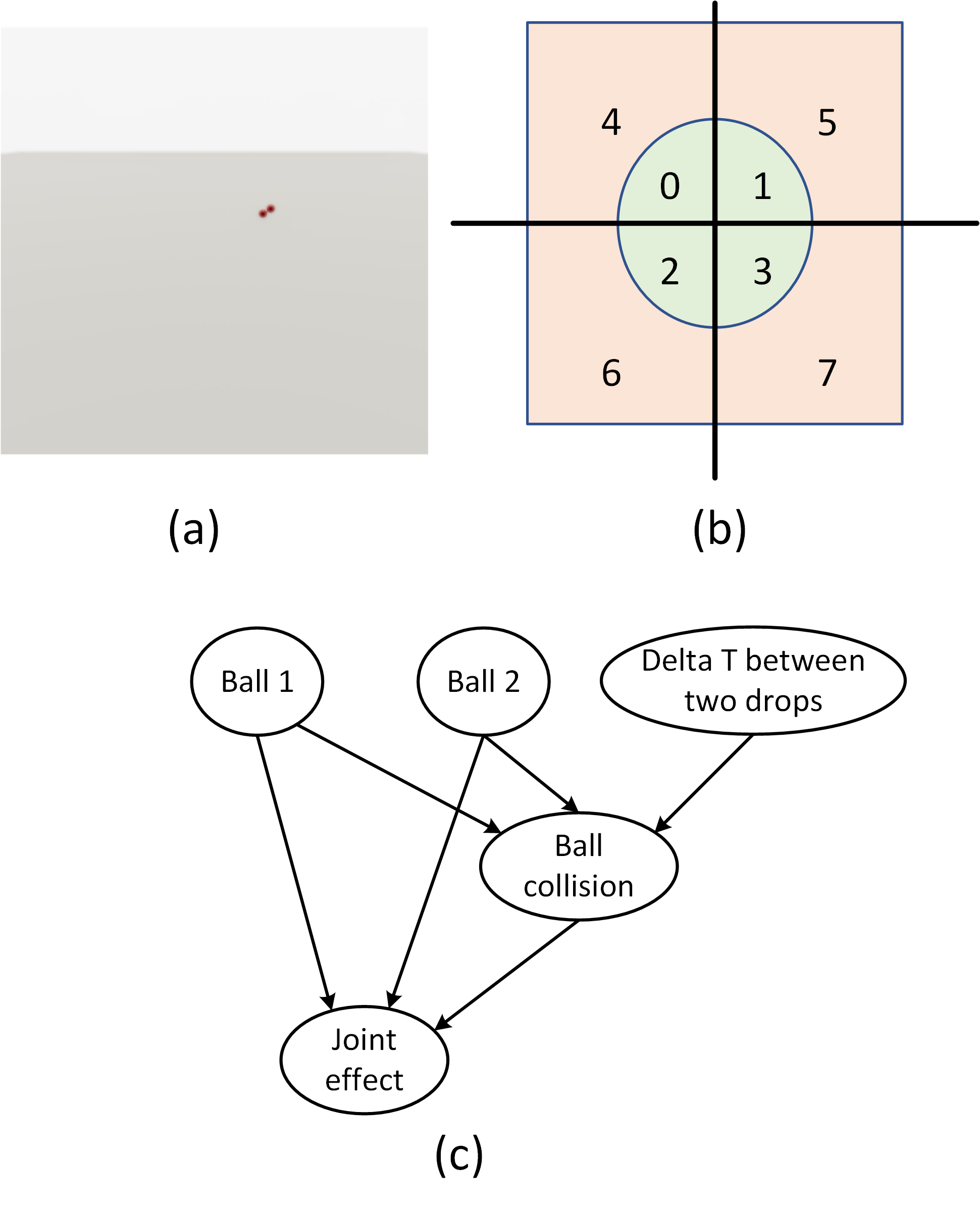}
\caption{(a) The simulation environment with possible collisions between two red balls. (b) Categorization of the final distance vector (as the effect) on a 2D plane. (c) A lumped causal graph of the causal structure of the experiment. }
\label{fig_exp_abc}
\end{figure}

To demonstrate the effectiveness of self-labeling in scenarios with complex causal structures, a simulation with multiple causes is designed and evaluated. TDW with PhysX engine is used to create the simulation environment \cite{tdw}. In this simulation, two balls are dropped onto a flat surface of size $150\times 150$ at randomized positions and times. The two balls will fall, potentially collide and interact, and eventually settle or reach the preset maximum simulation duration. The initial position of ball 2 is set to be higher than ball 1, and both are constrained in an area of size $20\times 20$ to produce a collision at a roughly 50\% rate. Collisions alter the balls' trajectories, complicating the causal structure and forcing the system to consider both causal paths. The final effect is a joint effect representing the distance vector from the final position of ball 1 to the final position of ball 2. The joint effect is discretized by categorizing the distance vector into 8 classes as described in \cref{fig_exp_abc}(b) depending on the vectors' angle and magnitude. Robustness to concept drift is tested by applying a perturbation in the form of wind, applied at a randomized bounded time to disturb balls' trajectories. The wind magnitude $wind$ is 0.5 by default and applied randomly to one ball. To penalize inaccurate interaction time inferences greater than the ground truth, the balls are instantiated with an initial horizontal velocity of 0.0025 before dropping.  
The chosen value of 0.0025 changes the initial distance of colliding balls by an average of 11\% and alters their joint behaviors, thus penalizing inaccurate interaction times.

\begin{figure}[!t]
\centering
\includegraphics[width=0.8\columnwidth]{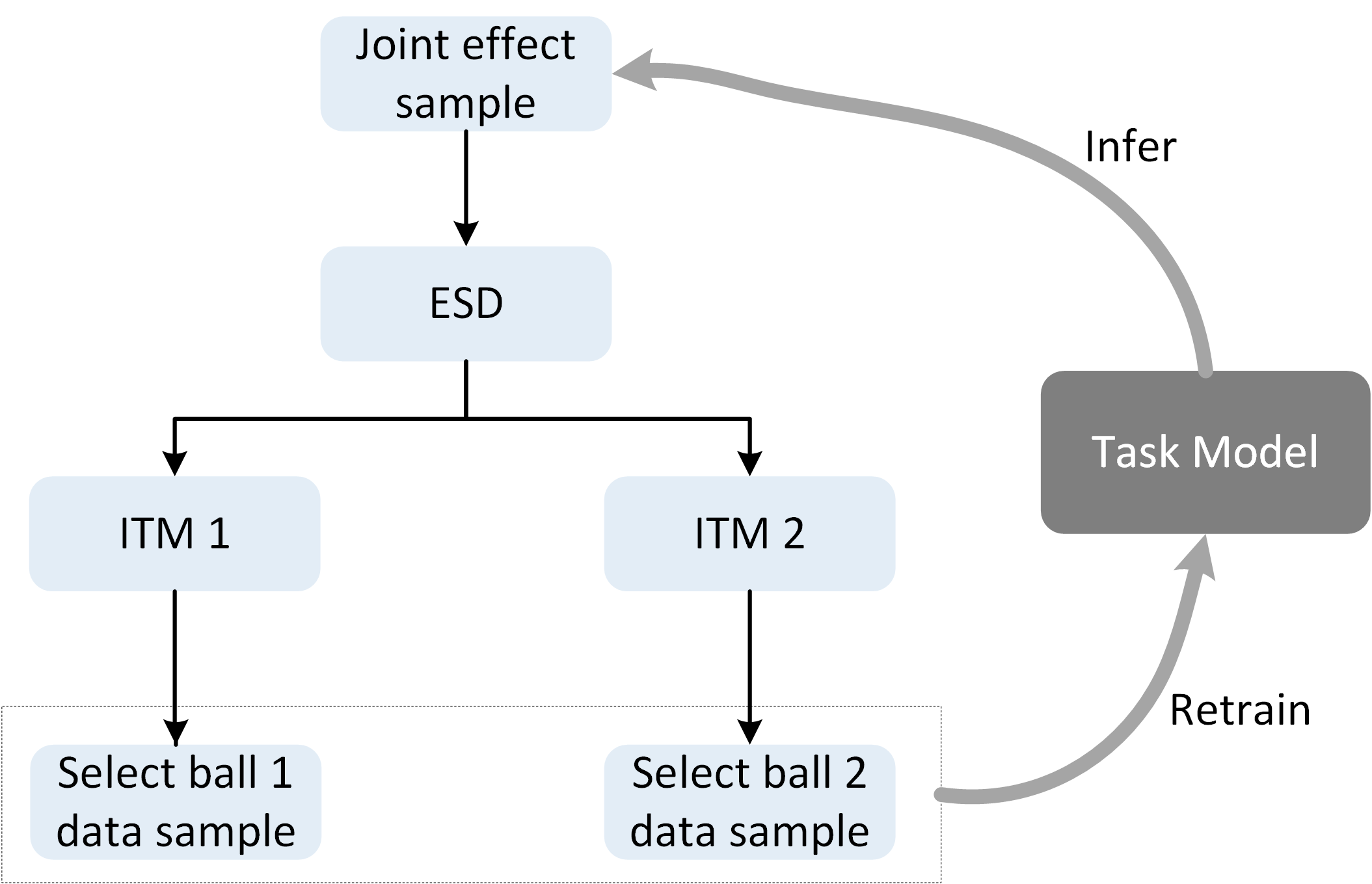}
\caption{A self-labeling workflow for the multivariate simulation.}
\label{ch3_fig_slb_itm}
\end{figure}

The causal graph for this simulation shown in \cref{fig_exp_abc}(c) contains two basic causal structures. The variables representing the initial position of ball 1, the initial position of ball 2, and the final effect form a collider structure. The addition of the possibility of collision creates a confounder structure within the graph. 

The objective of the task model is to use the two balls' initial properties to infer the class of the distance vector as the joint effect. Thus, the joint effect is used to self-label the cause events. As the cause states are transient, independent ITMs are required for each causal (cause-effect) pair. An interesting observation is that the observation of the ball collision is not necessary to self-label this scenario as the root causes of collision are observed. 
% This form of network contraction increases the causal depth of the self-labeling system. 
The holistic self-labeling workflow for this simulation is described in \cref{ch3_fig_slb_itm}. The two ITMs independently infer interaction times to select cause states from their respective data streams. The selected cause states are then combined as a self-labeled sample to retrain the task model.

\textbf{Dataset.} In total, 11700 class-balanced samples are used. The pre-training set has 600 samples. To simulate the incremental adaptiveness of learning, 360 samples are used per increment in the self-labeled dataset with 25 total increments. The test set is comprised of 1500 samples, and the validation set has 600 samples. The input for the task model is a 6-element vector comprised of the 3-$d$ and planar Euclidean distance of the two balls' initial positions, the 3-$d$ distance vector, and the interval between drops. 
The input features for the ITMs are vectors with 18 elements, including the 3-$d$ final positions and velocities of the two balls, their relative distance, the joint effect category, and the number of surface rebounds each ball experienced. The two data streams of monitoring each ball's properties before reaching the ground represent the two cause streams. The data stream of the joint effect after two balls reaching the ground represents the effect stream.

\begin{figure*}[!t]
\centering
\includegraphics[width=0.7\textwidth]{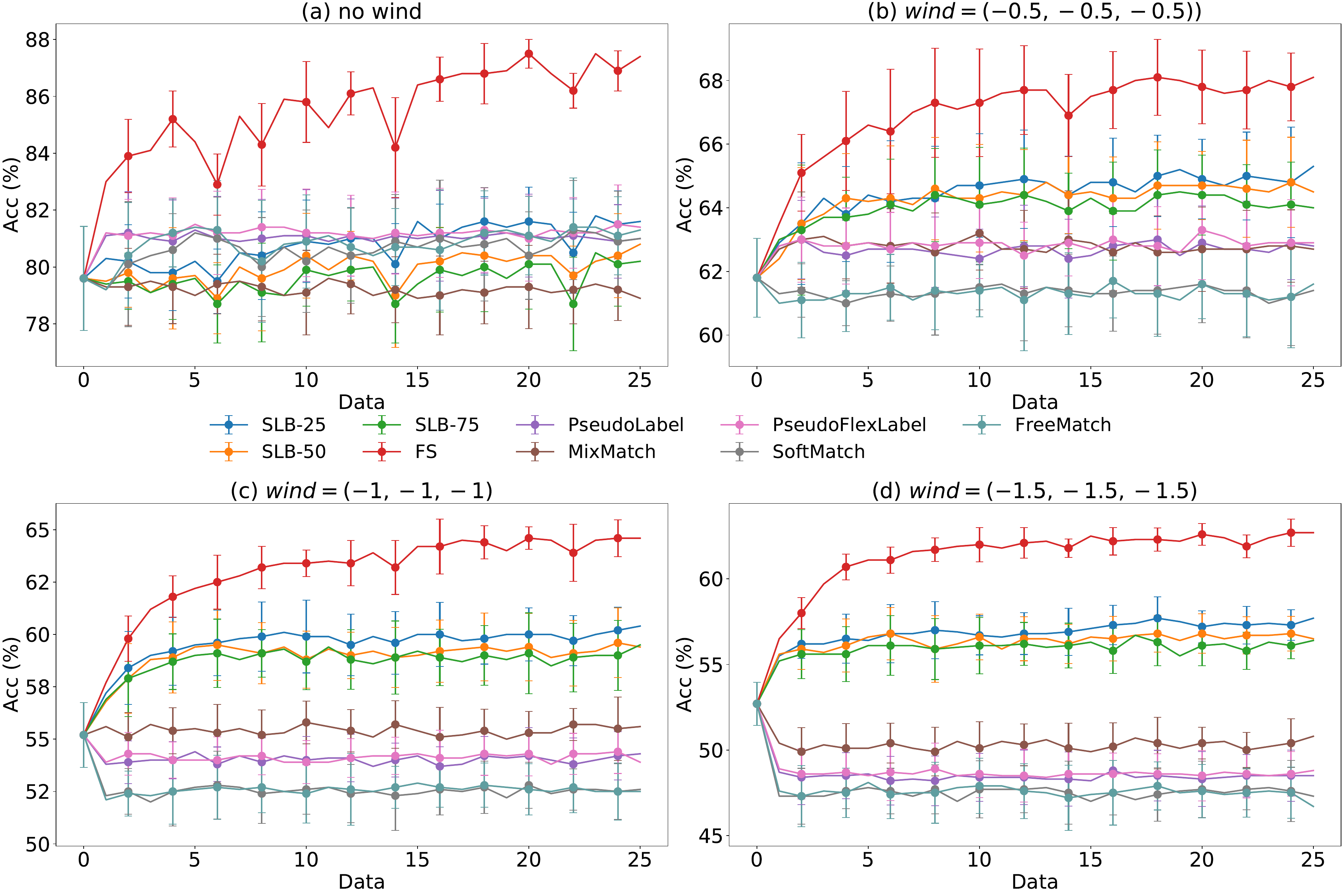}
\caption{Learning results with differing wind magnitudes. SLB results include varying ITM inaccuracy penalties (SLB-25, -50, and -75 represent penalties of 0.0025, 0.005, and 0.0075, respectively.) $X$ axis is the number of increments of the self-labeled datasets. Results at $X=0$ are from initial training on the pre-training set only. Error bars of standard deviation are plotted. For easier view, error bars are shown for every other increment.}
\label{ch3_fig_slb_result}
\end{figure*}

\begin{figure}[!t]
\centering
\includegraphics[width=1.0\columnwidth]{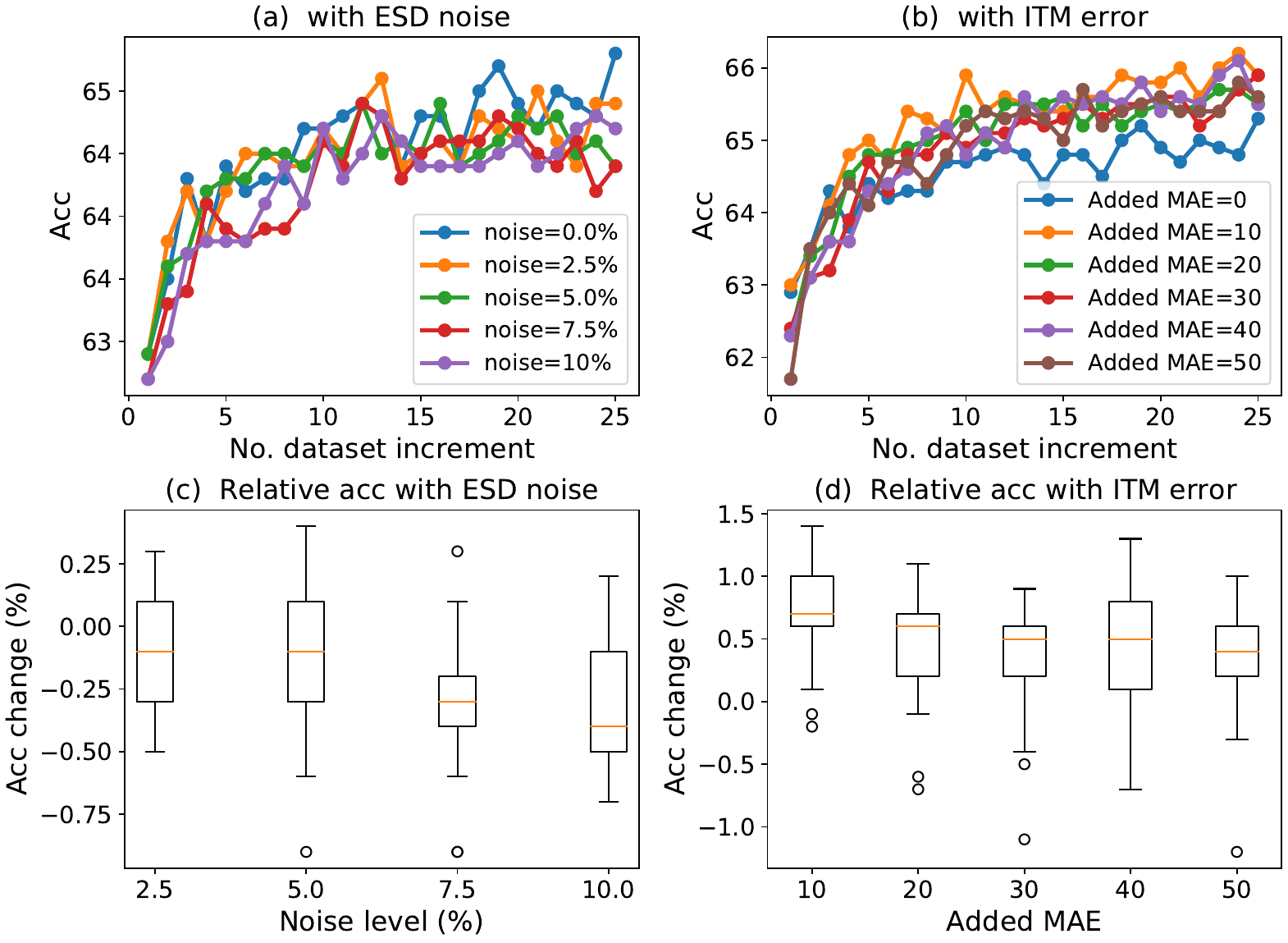}
\caption{Experiment results in the perturbed case with $wind=0.5$ across (a) label noise proportions and (b) added ITM MAE error. (c) and (d) represent the average accuracy shift from baseline upon induced ESD noise and ITM error, respectively, with error bars.}
\label{ch3_fig_noise}
\end{figure}

Nested k-fold validation is applied to performance evaluation as in \cite{ren2023slb}. 
The task model is a multi-layer perceptron (MLP) of size (32, 64, 128, 256, 128, 64, 32) with ReLU activation, a batch norm layer, and a dropout layer after each linear layer implemented using PyTorch and optimized by AdamW using a weight decay coefficient of 0.0005 and 0.001 learning rate. The batch size is 64 with 600 epochs. Two XGBoost models \cite{chen2016xgboost} optimizing for mean squared error loss are used as the ITMs for each cause data stream. ESDs use the categorization rule in \cref{fig_exp_abc}(b) to infer labels.
When evaluated on the perturbed dataset with wind magnitude $wind=0.5$, the R2 score for ITM 1 (ball 1) is 0.884, and its Mean Absolute Error (MAE) is 23. The R2 score of ITM 2 (ball 2) is 0.928, and its MAE is 17.7. 

\cref{ch3_fig_slb_result} shows training results for the self-labeling method, fully supervised learning, and five recent semi-supervised methods \cite{2013pseudo, 2019mixmatch, 2021flexmatch, wang2023freematch, chen2023softmatch}. Three wind magnitudes (0.5, 1.0, 1.5) are tested to evaluate concept drift resiliency. It can be observed that in the unperturbed case, self-labeling accuracy gradually increases, eventually outperforming other methods as it continues to learn via additional self-labeled samples. In the perturbed cases, self-labeling consistently outperforms other traditional SSL methods, further demonstrating its superiority in adapting to data shifts given complex causal structures. To further validate self-labeling effectiveness, we increase the penalty parameter to 0.005 and 0.0075 with results shown in \cref{ch3_fig_slb_result}. With an increased penalty, self-labeling performance is degraded but still outperforms traditional SSL in concept drift adaptation.

\subsection{Non-ideal ITM and ESD}

The impact of ESD inaccuracy is quantitatively tested using the multivariate simulation. We intentionally control ESD label noise by randomizing a portion of the ESD output to observe its effect on self-labeling performance. \cref{ch3_fig_noise} shows the experimental results with four levels of label noise in the perturbed case with $wind=0.5$. The result in \cref{ch3_fig_noise}(a) and (c) shows that ESD label noise both increases volatility and degrades self-labeling performance. The performance deterioration is mild, with an average decrease of 0.32\% at 10\% noise level, demonstrating the robustness of the self-labeling method against ESD inaccuracy.

Additionally, \cref{ch3_fig_noise}(b) shows the impact of ITM performance on task model accuracy.
This experiment quantifies the impact of ITM errors by modifying the baseline ITM output. While the baseline ITM output is not error-free in reality, we approximate it to be error-free for the purposes of this comparison. Additive MAE with random sign (positive or negative) is introduced to the baseline ITM error level, sampled from a Gaussian distribution with parameterized mean and variance. The variance is set as half of the mean which ranges from 0 to 50 with a step size of 10.
We can observe that with this random error added, SLB performance is slightly improved. In this perturbed case, as the ITMs are trained in the original domain, ITM inference is incongruent with perturbed interaction times and inherently deviates from the ground truth. The additive error can either improve or worsen this deviation. Its randomness functions as a compensation element in the self-labeling methodology, which can be beneficial to self-labeling performance, as shown in \cref{ch3_fig_noise}(d).
These experiments confirm the applicability of self-labeling in real-world applications with tolerance for imperfect ESD and ITM implementations.

\subsection{Cost analysis experiment}

\begin{figure}[!t]
\centering
\includegraphics[width=1.0\columnwidth]{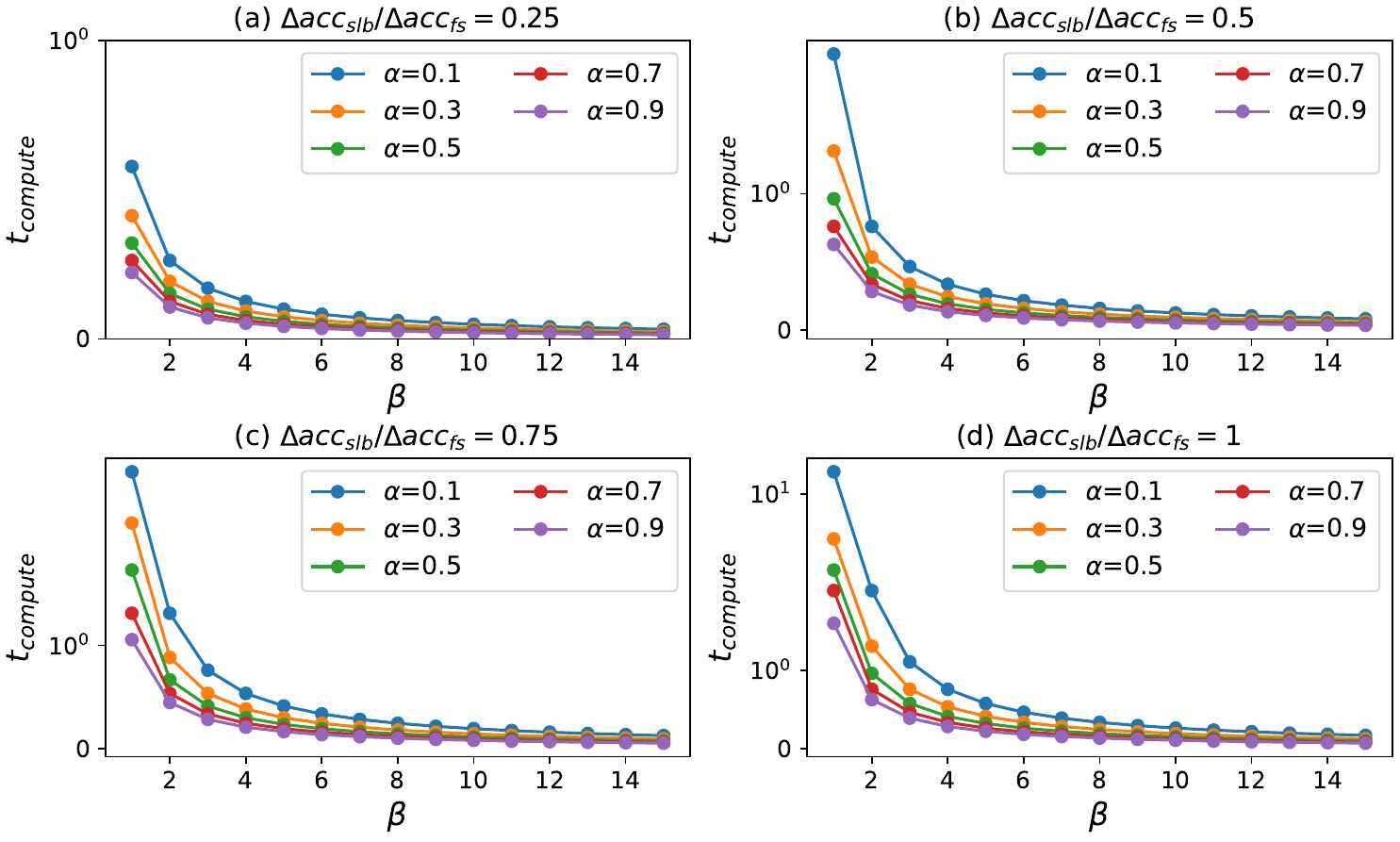}
\caption{Cost index analysis of $t_{compute}$ with respect to $\alpha$, $\beta$, and $\frac{\Delta acc_{slb}}{\Delta acc_{fs}}$.}
\label{fig_cost}
\end{figure}

Based on experimental figures and estimated values, we can perform a cost index analysis.
Amazon SageMaker Ground Truth\footnote{\url{https://aws.amazon.com/sagemaker/groundtruth/}} charges approximately $C_{m}=0.104$ per label, a sum of the price per each reviewed object (\$0.08) and the price of Named Entity Recognition (\$0.024). Reasonable estimations can be made for $\alpha$, $P$, and $r$ in \cref{eq_acc3}. Modern GPUs consume 200 to 450 Watts, and a nominal 400 W consumption (NVIDIA A100 \cite{a100}) is used in this study. The industrial electricity rate in the US is about \$0.05 to \$0.17 per kWh, and the average $r=\$ 0.09$ is used in the following analysis\footnote{\url{https://www.eia.gov/}}. 
Empirically, the ratio of inference to training time $\alpha$ is low, such that $0.1\leq \alpha \leq 1$. $t_{compute}$ is highly dependent on model size and $\frac{\Delta acc_{slb}}{\Delta acc_{fs}}$ is determined experimentally. \cref{fig_cost} plots the derived $t_{compute}$ against $\alpha$, $\beta$, and $\frac{\Delta acc_{slb}}{\Delta acc_{fs}}$. Using data from \cref{sec: exp_mul} and previous results in \cite{ren2023slb}, we can approximate that$\frac{\Delta acc_{slb}}{\Delta acc_{fs}}=0.5$ when $\beta = 1$. Empirically, we make the conservative estimate $\alpha=0.5$. Given the estimated parameters above, \cref{eq_acc3} can be solved to find that $t_{compute}\leq 1.3 h$, being that if the average training time per sample on a single GPU is less than 1.3 hours, $cost\_index_{slb} \geq cost\_index_{fs}$. In practice, this condition is easily satisfied. 

\cref{fig_cost} provides additional insights and comparisons. It is necessary to evaluate cost across $\beta$ as SLB may require more data than FS to achieve the same accuracy. In an extreme case where $\beta=15$, $\alpha=0.9$, $\frac{\Delta acc_{slb}}{\Delta acc_{fs}}=0.25$, we find $t_{compute} \leq 1 min$. For many mainstream image processing algorithms, \eg a benchmark by NVIDIA using A100 with ResNet50\footnote{\url{https://catalog.ngc.nvidia.com/orgs/nvidia/resources/resnet\_50\_v1\_5\_for\_tensorflow/performance}}, $t_{compute}=0.27s$ in training with 250 epoch, satisfying the  $t_{compute}$ requirement of 1 minute. 

Overall, with common $\alpha$ and $\beta$ values, SLB is generally cost-efficient relative to FS as long as both methods reach the desired accuracy for the application. This remains true as long as manual labeling costs far exceed the electricity costs per unit trained.

\subsection{Discussion}

\textbf{ITM and ESD error in real applications.}
This paper uses a simulation to validate that self-labeling has a high tolerance for ITM and ESD noise. 
Previous studies \cite{label_noise1, label_noise2, label_noise3} have shown that deep learning (DL) models are relatively robust to certain label noise levels.
While the ESD performance directly determines the label noise, the ITM is the input sampler for cause states in the cause data stream, selecting a period of samples in the cause data stream as the training input. The ITM error tolerance arises from the smoothness of state change transients in the real world. For example, a ball's movement and trajectory are smooth such that deviated interaction times can preserve the trend of motion for ML. However, as DL model tolerance for temporal shifts in input data has not yet been widely studied, this input nonideality appears in self-labeling and requires future study. Intuitively, ITM errors shift the sampling window, which may exclude moments with high information density or differentiating features, greatly hampering model performance. In addition, in practical applications, ESD noise has a second-order effect on ITM performance, as the ESD output may be included in the ITM input. This second-order effect can be studied in the future.

\textbf{Cost analysis.} It is evident that fully supervised learning has greater accuracy and resource consumption than methods on the unsupervised spectrum. This paper presents a cost index, including labor costs for data annotation, to compare the adaptive learning performance of FS and SLB. Despite traditional semi-supervised learning's advantage in resource consumption as calculated using equations in \cref{eq_cost_idx}, it is not included in this analysis as, experimentally, it has been found to achieve no observable or consistent accuracy improvement with increased data in the simulated experiment.
Outside of the two assumptions in \cref{sec: cost}, it is important to consider the accuracy figure achieved by self-labeling. The proposed metric does not account for the impact of accuracy in practical applications, where minor decreases in performance may result in a great impact on user experience.

\textbf{Domain knowledge modeling.}
This work relies on causality extracted from existing knowledge. Besides documented knowledge or domain experts, the potential of large language models (LLMs) reveals a rich knowledge base for extracting causality. Several pioneer works have demonstrated that LLMs are able to answer several types of causal questions \cite{zhang2023understanding, gpt_causal1, gpt_causal2, gpt_causal3}, while some work argues LLMs' ability of discovering novel causality \cite{gpt_causal0}. Using LLMs as the initial causal knowledge base for the proposed interactive causality-based self-labeling method will be an inspiring future work.

%% file: section6.tex
\section{Conclusion}
\label{sec: end}

This paper addresses several remaining questions in the interactive causality enabled self-labeling including multivariate causality application, robustness towards ITM and ESD error, and a cost and tradeoff analysis including manpower for self-labeling. The demonstration in this study further enhances the application values of self-labeling. More theoretical development of the interactive causality driven self-labeling is discussed as the future work in this direction.